\documentclass[sigconf]{acmart}
\usepackage{multirow}
\usepackage{colortbl} 
\usepackage{xcolor}
\usepackage{enumitem}

\copyrightyear{2023}
\acmYear{2023}
\setcopyright{acmlicensed}
\acmConference[CIKM '23] {Proceedings of the 32nd ACM International Conference on Information and Knowledge Management}{October 21--25, 2023}{Birmingham, United Kingdom.}
\acmBooktitle{Proceedings of the 32nd ACM International Conference on Information and Knowledge Management (CIKM '23), October 21--25, 2023, Birmingham, United Kingdom}
\acmPrice{15.00}
\acmISBN{979-8-4007-0124-5/23/10}
\acmDOI{10.1145/XXXXXX.XXXXXX}

\begin{document}
\title{Cross-heterogeneity Graph Few-shot Learning}

\author{Pengfei Ding}
\affiliation{%
 \institution{Macquarie University}
 \city{Sydney}
 \country{Australia}}
\email{pengfei.ding2@students.mq.edu.au}

\author{Yan Wang}
\authornote{Correspomding author}
\affiliation{%
 \institution{Macquarie University}
 \city{Sydney}
 \country{Australia}}
\email{yan.wang@mq.edu.au}

\author{Guanfeng Liu}
\affiliation{%
 \institution{Macquarie University}
 \city{Sydney}
 \country{Australia}}
\email{guanfeng.liu@mq.edu.au}

\begin{abstract}
In recent years, heterogeneous graph few-shot learning has been proposed to address the label sparsity issue in heterogeneous graphs (HGs), which contain various types of nodes and edges. The existing methods have achieved good performance by transferring generalized knowledge extracted from rich-labeled classes in source HG(s) to few-labeled classes in a target HG. However, these methods only consider the single-heterogeneity scenario where the source and target HGs share a fixed set of node/edge types, ignoring the more general scenario of cross-heterogeneity, where each HG can have a different and non-fixed set of node/edge types. To this end, we focus on the unexplored cross-heterogeneity scenario and propose a novel model for Cross-heterogeneity Graph Few-shot Learning, namely CGFL. In CGFL, we first extract meta-patterns to capture heterogeneous information and propose a multi-view heterogeneous graph neural network (MHGN) to learn meta-patterns across HGs. Then, we propose a score module to measure the informativeness of labeled samples and determine the transferability of each source HG. Finally, by integrating MHGN and the score module into a meta-learning mechanism, CGFL can effectively transfer generalized knowledge to predict new classes with few-labeled data. Extensive experiments on four real-world datasets have demonstrated the superior performance of CGFL over the state-of-the-art methods.
\end{abstract}



\maketitle
\section{Introduction}
Heterogeneous Graphs (HGs) consisting of diverse node types and diverse edge types, are pervasive in a wide variety of real-world systems, such as social networks, citation networks, and e-commerce networks \cite{shi2016survey}. When learning the representations of nodes and edges in HGs, label scarcity is a common issue due to the expense and difficulties of data annotation in practice \cite{yoon2022zeroshot}. To address this issue, heterogeneous graph few-shot learning (HGFL) has been developed recently, which combines few-shot learning methods \cite{zhang2022few} with heterogeneous graph neural networks \cite{wang2022survey}. The aim of HGFL is to transfer the generalized knowledge extracted from rich-labeled classes in source HG(s), to learn few-labeled classes in a target HG. HGFL has been successfully applied to various applications, such as few-shot node classification \cite{zhang2022few2}, few-shot link prediction \cite{chen2022meta}, and few-shot graph classification \cite{guo2021few}. 

\begin{figure}[t]
\centering
\scalebox{0.33}{\includegraphics{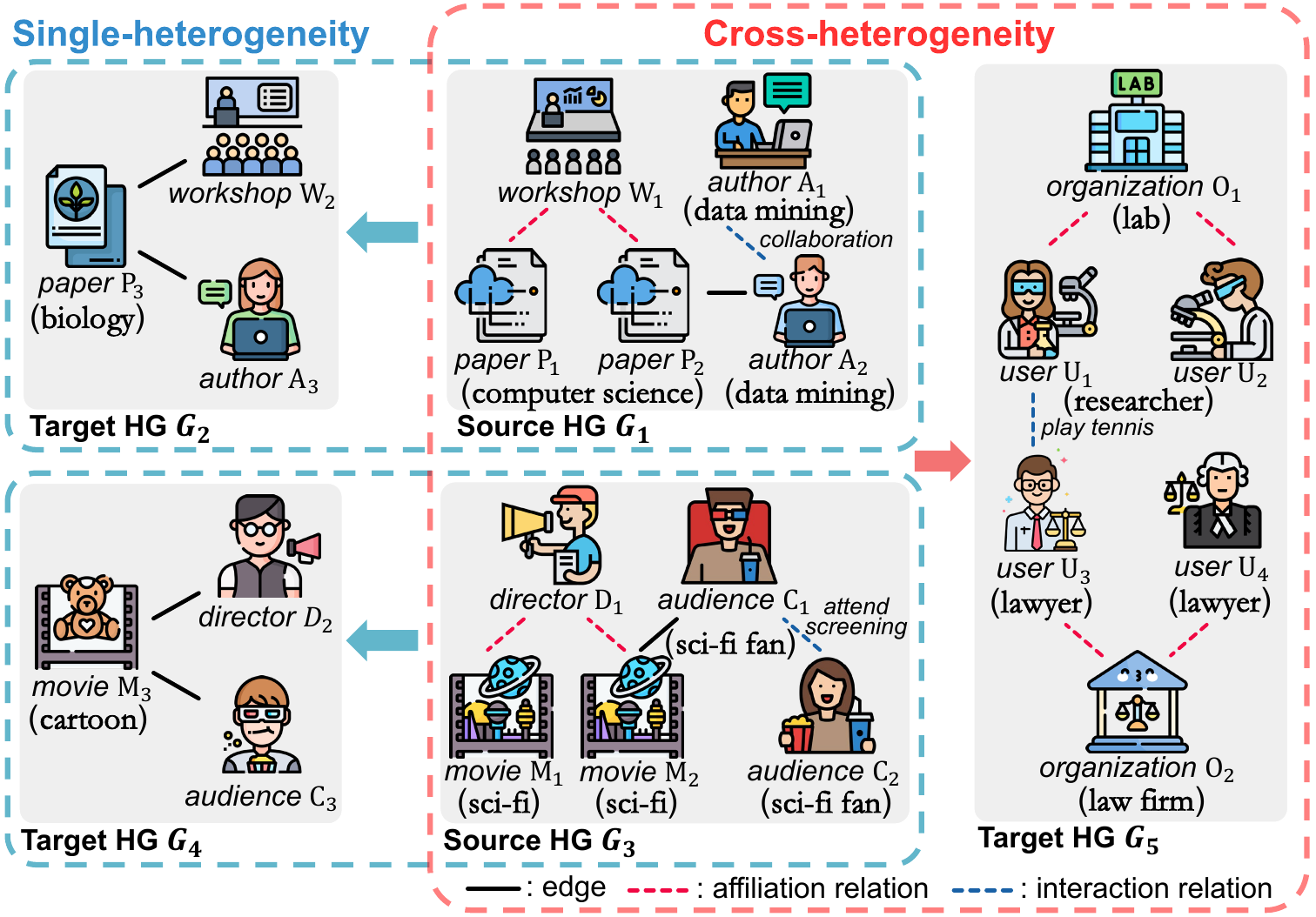}}
\caption{Scenarios of heterogeneous graph few-shot learning (HGFL).}
\label{eg1}
\end{figure}

Existing studies on HGFL mainly focus on the scenario of \emph{single-heterogeneity}, where the source and target HGs share a fixed set of node/edge types. In such a case, the knowledge of common node/edge types can be extracted from source HGs and transferred to the target HG to predict classes with few-labeled data. For instance, in Fig. \ref{eg1}, the source HG ${G}_1$ and target HG ${G}_2$ both contain node types of "paper", "workshop" and "author". The knowledge of the "paper" node type can be learned from ${G}_1$ and transferred to ${G}_2$ to predict the class of the "paper ${P}_3$" (\textit{i.e.}, \emph{biology}). Similarly, the knowledge of "movie" node type can be transferred from ${G}_3$ to ${G}_4$ to predict the class of the "movie ${M}_3$" (\textit{i.e.}, \emph{cartoon}).

However, in real-world scenarios, source HGs with rich-labeled classes may not always have the single-heterogeneity. In such a case, existing HGFL methods will be ineffective because they are unable to extract generalized knowledge across HGs through common node/edge types. For example, as shown in Fig. \ref{eg1}, if there are only two source HGs ${G}_1$ and ${G}_3$, existing HGFL methods cannot extract knowledge of "user" and "organization" from $\emph{G}_1$ and $\emph{G}_3$, thereby limiting their capability to predict the class of "users" in $\emph{G}_5$.

Based on the above discussion, a question arises: is it possible to extract generalized knowledge from other source HGs with different heterogeneities? In fact, different HGs may share relations that do not rely on specific node/edge types \cite{lu2019relation,ding2023few}. These shared relations can facilitate the extraction of generalized knowledge across HGs with different heterogeneities. One typical relation is the \emph{affiliation relation} \cite{lu2019relation,shi2020rhine}, which represents the association between nodes with individual properties and nodes with set properties. For instance, despite the disparate node types in ${G}_1$, ${G}_3$, and ${G}_5$, they all contain affiliation relations such as "paper-workshop", "movie-director", and "user-organization". By leveraging the similarities and patterns derived from the affiliation relations existing in source HGs ${G}_1$ and ${G}_3$ (\textit{e.g.}, "papers" published in the same "workshop" tend to have similar research topic labels, and "movies" directed by the same "director" tend to have similar theme labels), we can extract generalized knowledge that nodes with the same affiliation relation tend to have similar labels. This knowledge can be transferred to the target HG ${G}_5$ to classify "users" based on their work "organizations", \textit{e.g.}, $U_1$ and $U_2$ working at the same "organization" (\textit{i.e.}, \emph{lab}) may have similar professions (\textit{i.e.}, \emph{researchers}).

In addition to the affiliation relation, another important relation is the \emph{interaction relation} \cite{lu2019relation,shi2020rhine}, which represents the connection between nodes with individual properties. For instance, in Fig. \ref{eg1}, ${G}_1$, ${G}_3$, and ${G}_5$ all contain interaction relations such as "author-author", "audience-audience" and "user-user". Specifically, "authors" $A_1$ and $A_2$ in ${G}_1$ "collaborate" with each other, "audiences" $C_1$ and $C_2$ in ${G}_3$ "attend screenings" together, and $U_1$ and $U_3$ in ${G}_5$ "play tennis" together. By analyzing the labels of these nodes (\textit{e.g.}, both $A_1$ and $A_2$ focus on \emph{data mining}, while both $C_1$ and $C_2$ are \emph{sci-fi fans}), and the semantics of these connections (\textit{e.g.}, "collaborate" and "attend screenings" both represent forms of partnership), we can infer that nodes connected through interaction relations, which imply partnerships, are likely to share similar personal interests and preferences. This inference can be applied to the target HG ${G}_5$ to identify "users" (\textit{i.e.}, $U_1$ and $U_3$) who potentially share hobbies. 

The above-mentioned example illustrates the existence of common relations in HGs with different heterogeneities. These relations can be considered as general patterns that reflect the underlying connections among these HGs. We refer to these general patterns as \textit{meta-patterns}. Meta-patterns in real-world HGs may exhibit more complex structures and encompass a wider range of categories. Therefore, by further exploring various meta-patterns, we can discover generalized knowledge across HGs in the cross-heterogeneity scenario.



In light of the above discussion, the generalizability across HGs with different heterogeneities should be explored to handle a more general scenario of \emph{cross-heterogeneity graph few-shot learning}, namely, transferring generalized knowledge extracted from source HGs with multiple heterogeneities, to make predictions on the target HG with a different heterogeneity and few-labeled data. This is a novel problem, which, however, has the following key challenges.

\noindent\textbf{CH1.} \emph{How to normalize information from HGs with different heterogeneities to extract generalized knowledge implied in these HGs?} 
HGs with different heterogeneities typically exhibit significant variations in terms of node/edge features, node-node interactions, and graph structures. Therefore, in order to enable the extraction of generalized knowledge implied in these HGs, it is essential to develop a comprehensive framework that can normalize the heterogeneous information derived from HGs with various heterogeneities.

\noindent\textbf{CH2.} \emph{How to achieve effective knowledge transfer from source HGs to the target HG?} 
Since source HGs and the target HG differ in their structural similarities and heterogeneity relevance, not all source HGs contain sufficient generalized knowledge that can be transferred to the target HG. Additionally, even within the same HG, different samples may exhibit varying degrees of generalized knowledge due to their distinct surrounding heterogeneous environments. Therefore, to achieve effective knowledge transfer, it is crucial to select samples and HGs that encompass generalized knowledge across various HGs.


\noindent\textbf{CH3.} \emph{Given few-labeled samples from a new heterogeneity, how to effectively learn the classes associated with these samples?} In the target HG, since the few-labeled samples interact with various types of nodes and edges, each sample plays a distinct role in characterizing its respective class. However, measuring the importance of these samples from the new heterogeneity is challenging, because the knowledge of these interacted node/edge types cannot be transferred from source HGs. Consequently, to mitigate the influence of noise and outliers in these few-labeled samples, a robust and effective few-shot learning model is required.

To address the above three challenges, we propose a \textbf{C}ross-heterogeneity \textbf{G}raph \textbf{F}ew-shot \textbf{L}earning model, namely CGFL. In order to address CH1, we propose a general approach to extract meta-patterns across HGs with different heterogeneities, and adopt a novel multi-view heterogeneous graph neural network (MHGN) module to effectively generalize the information of these meta-patterns. To address CH2, we propose a three-level score module to evaluate (1) the transferability of source HGs, (2) the consistency of few-shot tasks, and (3) the informativeness of labeled samples. This module allows CGFL to perform hierarchical and preferential learning from the source HG data, facilitating the extraction of generalized knowledge in a stable manner and achieving effective knowledge transfer. To address CH3, we propose a novel meta-learning module that transfers the knowledge of measuring node informativeness from source HGs to the target HG. This module can effectively estimate the importance of few-labeled samples and generate highly robust class representations for accurate prediction in the target HG.

To the best of our knowledge, our work is the first to propose the novel problem of cross-heterogeneity graph few-shot learning. Our contributions can be summarized as follows:
\begin{itemize}
    \item We propose the CGFL model, which can learn transferable knowledge across HGs with multiple heterogeneities and adapt to predicting new classes with a different heterogeneity and few-labeled data;
    \item We propose a novel multi-view heterogeneous graph neural network module that can be generalized to HGs with different heterogeneities, and propose a novel three-level score module that can evaluate the source HG data to achieve effective knowledge transfer; 
    \item We conduct extensive experiments on four real-world datasets. Since our approach is the first to address this novel problem, we compare it against 12 representative and state-of-the-art baselines from four categories. The experimental results illustrate that CGFL outperforms the best-performing baselines by an average of 5.56\% in accuracy and 4.97\% in macro-F1 score.
\end{itemize}

\section{Related Work}
\noindent\textbf{Heterogeneous Graph Neural Networks (HGNNs).} 
HGNNs have shown promising results in learning representations of HGs. Some HGNNs directly model various types of nodes and edges. For example, HGT \cite{hu2020heterogeneous} calculates heterogeneous attention scores for 1-hop neighboring nodes w.r.t. edge types. Simple-HGN \cite{lv2021we} incorporates learnable edge type embeddings in edge attention, while HetSANN \cite{hong2020attention} employs a type-aware attention layer. SGAT \cite{lee2022sgat} extends GAT \cite{velickovic2017graph} to HGs using simplicial complexes and upper adjacencies. Other HGNNs focus on modeling meta-paths to extract hybrid semantics in HGs. For instance, HAN \cite{wang2019heterogeneous} designs node-level attention and semantic-level attention to hierarchically aggregate features from meta-path based neighbors. MAGNN \cite{fu2020magnn} further considers intermediate nodes along the meta-paths on the basis of HAN. However, existing HGNNs focus on learning specific node types and meta-paths within a single heterogeneity, which limits their ability to handle HGs with multiple heterogeneities.

\noindent\textbf{Graph Few-shot Learning.}
Most studies on graph few-shot learning focus on homogeneous graphs \cite{zhang2022few}. For example, Meta-GNN \cite{zhou2019meta} applies the MAML \cite{finn2017model} algorithm to address the low-resource learning problem on graphs. G-Meta \cite{huang2020graph} samples local subgraph surrounding the target node to transfer subgraph-specific information. TENT \cite{wang2022task} proposes to reduce the variance among tasks for generalization performance. Some recent studies have extended few-shot learning paradigms to heterogeneous graphs. For instance, HINFShot \cite{zhuang2021hinfshot} and HG-Meta \cite{zhang2022hg} target few-shot problems on a single citation network. CrossHG-Meta \cite{zhang2022few2} focuses on transferring knowledge across HGs with different graph structures but the same node/edge types. These approaches rely on the transfer of generalized knowledge through shared node/edge types between source and target HGs. Consequently, they are not applicable to the cross-heterogeneity scenario where the source and target HGs have completely different node/edge types.

Recently, MetaGS \cite{ding2023few} transfers knowledge between two HGs to predict semantic relations, it focuses on capturing comprehensive relationships between two nodes and requires the existence of semantic relations in the two HGs. Therefore, MetaGS cannot be generalized to deal with cross-heterogeneity few-shot problems.

\section{Preliminaries}
\noindent\textbf{Heterogeneous Graph.} A heterogeneous graph, denoted as ${G}$ = $(\mathcal{V}, \mathcal{E}, \phi, \psi)$, consists of a node set $\mathcal{V}$, an edge set $\mathcal{E}$, a node type mapping function $\phi$ : $\mathcal{V} \mapsto \mathcal{A}$, and an edge type mapping function $\psi: \mathcal{E} \mapsto \mathcal{R}$. $\mathcal{A}$ and $\mathcal{R}$ represent the set of node types and the set of edge types, respectively, where $|\mathcal{A}| + |\mathcal{R}| > 2$. The notations used in this paper are summarized in Table \ref{tab_nota}.

\noindent\textbf{Heterogeneity.} The heterogeneity of ${G}_{i}$, denoted as $\mathcal{H}_{i}$ = $(\mathcal{A}_{i}, \mathcal{R}_{i})$, contains the node type set $\mathcal{A}_{i}$ and the edge type set $\mathcal{R}_{i}$. ${G}_{i}$ and ${G}_{j}$ have {different heterogeneities} if $\mathcal{A}_{i} \cap \mathcal{A}_{j}$ = $\emptyset$ and $\mathcal{R}_{i} \cap \mathcal{R}_{j}$ = $\emptyset$.

\noindent\textbf{Problem Formulation.} We focus on the problem of few-shot node classification across HGs with different heterogeneities. To mimic the few-shot scenario, the data of each heterogeneity $\mathcal{H}_{i}$ is considered as a collection of ${m}$ few-shot tasks ${T}_{i}$=$\{\tau_1, \tau_2, \ldots, \tau_{m}\}$. 
\begin{itemize}[leftmargin=*]
    \item \textbf{Input:} Source HGs $\mathcal{G}_\emph{src}$ = $\{{G}_1, {G}_2, \ldots, {G}_{n}\}$ and their sets of few-shot tasks $\mathcal{S}$=$\{{T}_1, {T}_2, \ldots, {T}_{n}\}$. Target HG ${G}_{t}$ with a set of few-shot tasks $\mathcal{T}$. Each graph in $\mathcal{G}_\emph{src}$ has a different heterogeneity to ${G}_{t}$.
    \item \textbf{Output:} A model with good generalization ability to few-shot tasks on the target HG (\textit{i.e.}, $\mathcal{T}$), after training with few-shot tasks on source HGs (\textit{i.e.}, $\mathcal{S}$).
\end{itemize}

\noindent\textbf{Few-shot Task Construction.} Few-shot tasks ${T}$=$\{\tau_1, \tau_2, \ldots,$ $\tau_{m}\}$ in ${G}$ are constructed as follows. Firstly, under the ${N}$-way ${K}$-shot setting, each task $\tau$ = $(\tau_\emph{spt}, \tau_\emph{qry})$ is sampled by randomly choosing ${N}$ different classes ${C}_{\tau}$=$\{c_1, c_2, \ldots, c_{N}\}$ from the label space ${C}$. Specifically, the support set $\tau_\emph{spt}$ = $\{\tau_{{c}_1}, \tau_{{c}_2}, \ldots, $ $\tau_{{c}_{N}}\}$ is created by sampling ${K}$ labeled nodes per class, \textit{i.e.}, $\tau_{{c}_{i}}$ = $\{({v}_1, {c}_{i}), ({v}_2, {c}_{i}), \ldots, ({v}_{K}, {c}_{i})\}$. The query set $\tau_\emph{qry}$ = $\{\tilde{\tau}_{{c}_1}, \tilde{\tau}_{{c}_2}, \ldots, \tilde{\tau}_{{c}_{N}}\}$ is formed from the remaining data of each class, where $\tilde{\tau}_{{c}_{i}}$ = $\{(\tilde{{v}}_1, {c}_{i}), (\tilde{{v}}_2, {c}_{i}), \ldots, (\tilde{{v}}_\emph{K}, {c}_{i})\}$. After sufficient training iterations over $\mathcal{S}$ with the proposed methodology, the obtained model is expected to conduct ${N}$-way classification in $\mathcal{T}$ given only ${K}$ labeled examples per class.

\begin{table}[]
\caption{Notations used in this paper.}
\label{tab_nota}
\centering
\begin{tabular}{cc}
\toprule[1pt]
\textbf{Notation}                      & \textbf{Explanation}                    \\ \midrule[0.5pt]
${G}$              & heterogeneous graph      \\
$\mathcal{V}$         & set of nodes          \\
$\mathcal{E}$         & set of edges          \\
$\phi$         & node type mapping function  \\ 
$\psi$         & edge type mapping function          \\
$\mathcal{A}$         & set of node types          \\
$\mathcal{R}$         & set of edge types          \\
$\mathcal{H}$         & single heterogeneity       \\
${T}_i$         & set of few-shot tasks on $\mathcal{H}_i$\\
$\mathcal{G}_\emph{src}$         & set of source HGs       \\
${G_{t}}$         & target HG       \\
$\mathcal{S}$         & set of few-shot tasks on source HGs  \\
$\mathcal{T}$         & set of few-shot tasks on the target HG  \\
$\tau$ = $(\tau_\emph{spt}, \tau_\emph{qry})$  & single few-shot task\\
$\mathcal{P}$         & set of meta-patterns \\
$\mathcal{I}$         & set of instances matching meta-patterns in  $\mathcal{P}$\\
$\mathbf{proto}_c$         & prototype embedding of class $c$ \\
\bottomrule[1pt]
\end{tabular}
\end{table}

\section{The Proposed CGFL}
\begin{figure*}[t]
\centering
\scalebox{0.275}{\includegraphics{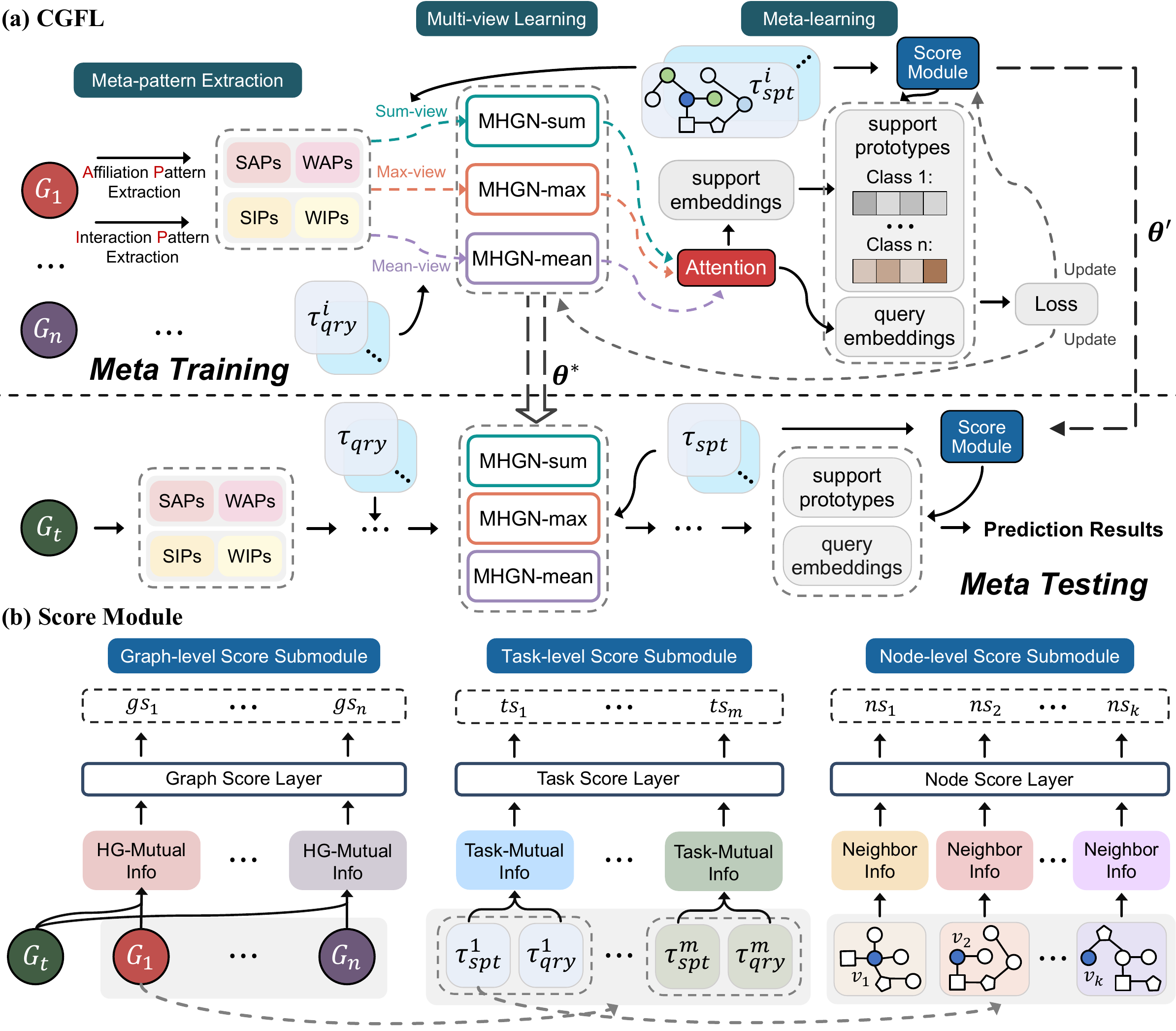}}
\caption{(a) The overall architecture of CGFL. (b) The three-level score module.}
\label{model}
\end{figure*}

Fig. \ref{model}(a) shows the structure of our proposed CGFL model, which contains three main steps: (1) \emph{Meta-pattern Extraction}: \emph{meta-patterns} are extracted and classified into several general categories to capture heterogeneous information across HGs; (2) \emph{Multi-view Learning}: a new multi-view heterogeneous graph neural network (MHGN) module is proposed to aggregate meta-pattern information from three general views: \emph{sum}, \emph{max} and \emph{mean}; (3) \emph{Meta-learning}: In the meta-training process, a three-level score module is proposed to assess the transferability of source HGs, the consistency of few-shot tasks, and the informativeness of labeled nodes. In the meta-testing process, the node-level score submodule is generalized to the target HG to evaluate the importance of each few-labeled node. Overall, CGFL can be well initialized from few-shot tasks in source HGs and adapt to the target HG, enabling accurate prediction of few-labeled classes from a different heterogeneity. 

\subsection{Meta-pattern Extraction}
In existing methods, the extraction of heterogeneous information from HGs typically relies on the utilization of predefined \emph{patterns} referred to as meta-paths \cite{sun2011pathsim} or meta-graphs \cite{fang2019metagraph}. These patterns consist of diverse node types and have specific semantics, such as "author-paper-author" indicating author collaboration. By matching the HG with multiple pre-defined patterns, various types of heterogeneous information embedded in the HG can be extracted for further analysis. However, the process of mining patterns for a single HG typically requires domain expertise \cite{chen2017task}, thereby presenting challenges in extracting general patterns in the cross-heterogeneity scenario. On the one hand, mining useful patterns for all source HGs is impractical due to the extensive domain knowledge required. On the other hand, mining valuable patterns in the target HG becomes arduous due to limited knowledge regarding the new heterogeneity. 

To tackle this challenge, we propose a novel method for extracting heterogeneous information across HGs automatically. Specifically, we first extract important patterns in HGs and form \textit{meta-patterns}. These meta-patterns are then classified into four general categories, enabling the extraction of generalized knowledge across HGs with different heterogeneities.

\noindent\textbf{Procedure for Extracting Meta-patterns.} For a single HG ${G}$, we propose a random walk-based approach to capture prevalent local structures and form meta-patterns: (1) Starting from each node in ${G}$, ${N}_\emph{path}$ paths with a length of ${l}$ are randomly chosen for each node, and these paths are aggregated into a set $\mathbf{S}_\emph{path}$. (2) For every node type $\Phi_{i} \in\mathcal{A}$ ($\Phi_{i}$ denotes the $i$-th node type), subpaths that connect nodes with the same node type $\Phi_{i}$ are extracted from $\mathbf{S}_\emph{path}$, forming the set $\mathbf{S}_\emph{path}^{i}$. (3) $\mathbf{S}_\emph{path}^{i}$ is then partitioned into distinct groups based on the type patterns of the subpaths (\textit{e.g.}, subpaths with the "author-paper-author" pattern and the "author-venue-author" pattern are assigned to different groups). (4) Meta-patterns for $\Phi_{i}$ are identified by selecting the patterns with the top-${K}_\emph{mp}$ highest counts. Hence, the meta-patterns in ${G}$ can be represented as $\mathcal{P}$ = $\{\mathcal{P}_i|\Phi_{i} \in\mathcal{A}\}$.

By identifying and summarizing patterns in HGs to form meta-patterns, we can effectively compress a large volume of diverse heterogeneous information into a more concise format. In addition, applying a uniform approach to deal with various HGs can facilitate extracting generalized knowledge from these HGs.


\noindent\textbf{Categorization for Meta-patterns.}
Our categorization approach firstly divides meta-patterns into two main categories: \emph{affiliation patterns} (APs) and \emph{interaction patterns} (IPs). Within each category, we further classify the patterns into two subcategories based on the strength of the relationship they represent. For instance, APs can be further categorized as either \emph{strong affiliation patterns} (SAPs) or \emph{weak affiliation patterns} (WAPs). This categorization is inspired by existing studies that adopt two general types of relations to classify various relationships in HGs \cite{lu2019relation,shi2020rhine}: the \emph{affiliation relation} between nodes with individual properties and nodes with set properties (\textit{e.g.}, "paper-venue"), and the \emph{interaction relation} between nodes with individual properties (\textit{e.g.}, "author-author"). 

However, these relations can only capture generalized node-node connections and are incapable of capturing generalized graph structures across HGs. Moreover, in the cross-heterogeneity scenario, each relation category may contain diverse information that requires further classification. To overcome these limitations, CGFL extends the two types of relations from node-level to a more intricate meta-pattern level and provides a more detailed categorization of meta-patterns. Specifically, we first categorize meta-patterns into APs and IPs as follows.
\begin{gather}
{D}_{mp}({p}) = \max(\{{D}(\Phi_{i}, \Phi_{j})|(\Phi_{i}, \Phi_{j}) \in {p}\}), 
 \\
 {D}(\Phi_{i}, \Phi_{j}) = \frac{\max({d}_{\Phi_{i}}, {d}_{\Phi_{j}})}{\min({d}_{\Phi_{i}}, {d}_{\Phi_{j}})},
\end{gather}
where ${p}$ is a single meta-pattern, $\emph{d}_{\Phi_{i}}$ and ${d}_{\Phi_{j}}$ are average degrees of node types $\Phi_{i}$ and $\Phi_{j}$ respectively. ${D}(\Phi_{i}, \Phi_{j})$ determines the type of relationship between $\Phi_{i}$ and $\Phi_{j}$ \cite{shi2020rhine}. A large value of ${D}(\cdot)$ indicates an affiliation relation, while a low value suggests an interaction relation. Then, we set a threshold $\theta_\emph{mp}$ to classify meta-patterns, with those having ${D}_\emph{mp}(\cdot) \geq \theta_\emph{mp}$ classified as affiliation patterns and those with ${D}_\emph{mp}(\cdot) < \theta_\emph{mp}$ classified as interaction patterns. The classification is based on whether the meta-pattern contains the affiliation relation or not. $\theta_\emph{mp}$ is typically set to 10, which is in line with previous studies on the classification of affiliation relations and interaction relations \cite{shi2020rhine}.

Next, we categorize affiliation patterns (APs) and interaction patterns (IPs) based on the strength of the relationships they represent. For APs, we analyze whether the pattern is symmetric, as symmetrical patterns typically indicate stronger relationships \cite{sun2011pathsim}. For instance, the symmetric AP "user-company-user" indicates that two "users" are connected through the same type of affiliation relation (\textit{i.e.}, "user-company"), suggesting that these two "users" are likely to be colleagues working in the same "company". Conversely, the asymmetric AP "user-company-item-user" indicates that two "users" are connected through different types of affiliation relations (\textit{i.e.}, "user-company" and "company-item-user"). In this case, one "user" may be an employee of the "company" while the other could be a customer of its products. These two "users" may have distinct roles, which weakens the relationship between them compared to the "user-company-user" pattern. Consequently, a symmetrical AP is classified as a \textit{strong affiliation pattern} (SAP), whereas an asymmetrical AP is categorized as a \textit{weak affiliation pattern} (WAP).

Regarding IPs, since each IP does not contain affiliation relations, nodes in IPs interact with each other on an equal basis. Nodes exhibit strong interaction relationships when they are closely connected, either directly or through a few common neighbors. Therefore, IPs with shorter lengths represent strong interaction relationships and are considered \emph{strong interaction patterns} (SIPs), while IPs with longer distances do not indicate a close interaction relationship and are classified as \emph{weak interaction patterns} (WIPs). We set a length threshold $\theta_\emph{lp}$ to specify SIPs and WIPs. 

Categorizing APs and IPs allows for a more comprehensive study of diverse meta-patterns. This categorization helps discover latent correlations within meta-patterns from each category, thereby facilitating the extraction of generalized knowledge. The categorization of meta-patterns yields four distinct groups: $\mathcal{P}^\emph{SAP}$, $\mathcal{P}^\emph{WAP}$, $\mathcal{P}^\emph{SIP}$, and $\mathcal{P}^\emph{WIP}$, each with its own set of instances, e.g., $\mathcal{I}^{\emph{SAP}}$.

\subsection{Multi-view Learning}
Each meta-pattern category can be represented by aggregating the information from its individual instances. However, directly aggregating meta-pattern instances may result in a loss of valuable information due to the different structures and node types in each category. To address this issue, we propose a multi-view mechanism inspired by widely used graph aggregation methods \cite{xu2018powerful}. This mechanism captures information from three perspectives: the \emph{sum-view}, \emph{max-view}, and \emph{mean-view}. These views have been proven to be general enough to encompass the capabilities of various graph aggregators \cite{xu2018powerful}, thereby enabling CGFL to extract comprehensive and generalized knowledge across HGs.
\begin{itemize}[leftmargin=*]
    \item In the sum-view, we capture the complete information of instances for each meta-pattern category, \textit{i.e.}, $\mathcal{I}^{\emph{sum-SIP}}$ = $\mathcal{I}^{\emph{SIP}}$, here we use SIP as an example for the three views.
    \item In the max-view, we focus on extracting information related to the most prevalent meta-pattern within each category:
    \begin{equation}
    \label{max_ins}
    \mathcal{I}^{\emph{max-SIP}} = \{{I}| {I} \in \mathcal{I}^{\emph{SIP}}, \mathcal{P}({I}) = {p}^\emph{SIP}_\emph{N-max}\},
    \end{equation}
    where $\mathcal{P}({I})$ represents the meta-pattern associated with instance ${I}$, ${p}^\emph{SIP}_\emph{N-max}$ is the meta-pattern in $\mathcal{P}^{\emph{SIP}}$ that has the maximum number of instances.
    \item In the mean-view, we capture the distributions of all meta-patterns by averagely sampling ${N}_\emph{mean}$ instances for each meta-pattern ${p} \in \mathcal{P}^{\emph{SIP}}$ and forming the set $\mathcal{I}^{\emph{mean-SIP}}$. 
\end{itemize}
Thus, we can create instance sets for the three views: $\mathcal{I}^{\emph{sum}}$, $\mathcal{I}^{\emph{max}}$ and $\mathcal{I}^{\emph{mean}}$. Each set consists of instances corresponding to different meta-pattern categories, \textit{e.g.}, $\mathcal{I}^{\emph{sum}}$=$\{\mathcal{I}^{\emph{sum-SAP}}$, $\mathcal{I}^{\emph{sum-WAP}}$, $\mathcal{I}^{\emph{sum-SIP}}$, $\mathcal{I}^{\emph{sum-WIP}}$\}. By leveraging these views to learn meta-patterns from different categories, we can extract comprehensive and generalized knowledge across HGs. Next, we adopt the following three functions to aggregate the information of meta-patterns and compute node representations.

\noindent\textbf{Meta-pattern Instance Aggregation.}
To begin with, we perform the aggregation of instances belonging to the same meta-pattern category and the same view (\textit{e.g.}, $\mathcal{I}^{\emph{sum-SIP}}$). Let $\mathcal{I}_{v}^{\emph{sum-SIP}}$ represent a set of instances that start from node $v$ and are extracted from $\mathcal{I}^{\emph{sum-SIP}}$. An instance encoder is employed to transform the node features of each instance into a single vector:
\begin{equation}
\mathbf{h}_{v\emph{-}i}^{\emph{sum-SIP}}=f^\emph{SIP}_{\theta}(\{\mathbf{{x}}_{u}|{u}\in{I}_{v\emph{-}i}^{\emph{sum-SIP}}\}),
\end{equation}
where $\mathbf{x}_\emph{u}$ is the feature of node $\emph{u}$, ${I}_{v\emph{-}i}^{\emph{sum-SIP}}$ represents the ${i}$-th instance in $\mathcal{I}_{v}^{\emph{sum-SIP}}$, and $f^\emph{SIP}_{\theta}(\cdot)$ denotes the shared encoder function for SIP instances. Since each instance contributes differently to the representation of the target node, we learn the importance weight for each instance and utilize a multi-head attention mechanism to calculate weighted sums of the instance representations:
\begin{gather}
\alpha^{\emph{sum-SIP}}_{v\emph{-}i} = \frac{\exp\left(\sigma\left(\mathbf{a}_\emph{sum-SIP}^{\mathsf{T}}\cdot\left[\mathbf{{x}}_{v} \| \mathbf{h}_{v\emph{-}i}^{\emph{sum-SIP}}\right] \right)\right)}{\sum\nolimits^{|\mathcal{I}_{v}^{\emph{sum-SIP}}|}_{{j}=1}\exp\left(\sigma\left(\mathbf{a}_\emph{sum-SIP}^{\mathsf{T}}\cdot\left[\mathbf{{x}}_{v} \| \mathbf{h}_{v\emph{-}j}^{\emph{sum-SIP}}\right] \right)\right)},
 \\
\mathbf{h}_{v}^{\emph{sum-SIP}}=\mathop{\parallel}^{{K}_\emph{att}}_{{k}=1}\sigma\left(\sum\nolimits^{|\mathcal{I}_{v}^{\emph{sum-SIP}}|}_{{i}=1} \alpha^{\emph{sum-SIP}}_{{v\emph{-}i, k}} \cdot \mathbf{h}_{v\emph{-}i}^{\emph{sum-SIP}}\right), 
\end{gather}
where $\alpha^{\emph{sum-SIP}}_{v\emph{-}i, k}$ denotes the importance of ${i}$-th instance at the ${k}$-th attention head, $\|$ denotes the concatenation operation, $\mathbf{a}_\emph{sum-SIP}$ represents the trainable attention parameter shared among SIP instances obtained from the sum-view, ${K}_\emph{att}$ corresponds to the number of attention heads, and $\sigma(\cdot)$ denotes an activation function.

\noindent\textbf{Meta-pattern Category Aggregation.} 
Next, we aggregate the information from the four meta-pattern categories under the same view. Specifically, we adopt an attention mechanism to aggregate information from these meta-pattern categories:
\begin{gather}
\label{test}
{w}_\emph{sum-SIP} = \sigma\left(\left({\mathbf{W}_\emph{sum}\cdot \mathbf{h}_{v}^{\emph{sum-SIP}} + \mathbf{b}_\emph{sum}}\right)\cdot\mathbf{a}_\emph{sum}\right), \\
\beta_\emph{sum-SIP} = \exp({w}_\emph{sum-SIP})/\sum\nolimits_{\emph{i}\in \mathbf{S}_\emph{mp}}\exp({w}_\emph{sum-i}),\\ 
\mathbf{h}^{\emph{sum}}_{v} = \sum\nolimits_{{i}\in \mathbf{S}_\emph{mp}}\beta_\emph{sum-i}\cdot \mathbf{h}_{v}^{\emph{sum-i}},
\end{gather}
where $\mathbf{S}_\emph{mp}$ = $\{\emph{SAP, WAP, SIP, WIP}\}$ is the set of meta-pattern categories, $\mathbf{W}_\emph{sum}$ and $\mathbf{b}_\emph{sum}$ are trainable parameters, $\mathbf{a}_\emph{sum}$ denotes the shared attention parameter for meta-pattern categories in the sum-view.

\noindent\textbf{View Aggregation.}
Finally, we aggregate the information from the three views to obtain the final embedding:
\begin{gather}
z_\emph{sum} = \sigma\left(\left({\mathbf{W}_\emph{view}\cdot \mathbf{h}_{v}^{\emph{sum}} + \mathbf{b}_\emph{view}}\right)\cdot\mathbf{a}_\emph{view}\right), \\
\gamma_\emph{sum} = \exp({z}_\emph{sum})/\sum\nolimits_{{i}\in \mathbf{S}_\emph{view}}\exp({z}_{i}),\\ 
\mathbf{h}_{v} = \sum\nolimits_{{i}\in \mathbf{S}_\emph{view}}\gamma_{i}\cdot \mathbf{h}_{v}^{{i}},
\end{gather}
where $\mathbf{S}_\emph{view}$ = $\{\emph{sum, max, mean}\}$ is the set of views, $\mathbf{W}_\emph{view}$ and $\mathbf{b}_\emph{view}$ are trainable parameters, and $\mathbf{a}_\emph{view}$ denotes a view level attention parameter. 

\subsection{Meta-learning}
The meta-learning module consists of two steps: meta-training and meta-testing. During meta-training, to achieve effective knowledge transfer from source HGs, we propose a three-level score module (as shown in Fig. \ref{model}(b)) that evaluates the source HGs' data from three perspectives: HG transferability, task consistency, and node informativeness. Subsequently, we adopt a prototypical network to transfer the generalized knowledge from source HGs to the target HG. In meta-testing, the node-level score submodule, which has been learned from the source HGs, is adapted to the target HG. This adaptation enables the evaluation of the importance of few-labeled nodes in the target HG, thereby facilitating the generation of more stable representations for classes in the target HG.

\noindent\textbf{Graph-level Score Submodule.}
This submodule is proposed to evaluate the transferability of each source HG. By comparing the similarity of heterogeneous information between the source HG and the target HG, we determine whether the source HG contains sufficient generalized knowledge for effective knowledge transfer. Specifically, we calculate the transferability of each source HG by comparing the meta-pattern distributions between this source HG and the target HG. Since the mean-view in the multi-view learning module captures the distributions of all meta-patterns, we utilize the vector $\mathbf{h}_v^\emph{mean}$ output by the mean-view (\textit{cf}. Eq. 9) to compute the representation of the meta-pattern distribution for each HG:
\begin{equation}
\mathbf{h}_{G_i} = \emph{mean}(\{\mathbf{h}_v^\emph{mean}|v\in\tau, \tau\in T_i\}),
\end{equation}
where $\tau$ is an $N$-way $K$-shot task, $T_i$ is the set of tasks for $i$-th source HG $G_i$, and \emph{mean}$(\cdot)$ denotes the averaging operation. For the target HG $G_t$, we randomly select $N\times K \times |T_i|$ nodes and input them into the current model to obtain $h_v^\emph{mean}$ and form $h_{G_t}$ using the \emph{mean}$(\cdot)$ operation. Note that this process is solely for obtaining representations of different HGs in the same vector space, and therefore the gradients during the vector acquisition process are not optimized. Next, we concatenate the representations of each source HG and the target HG and adopt softmax to compute the graph-level score for each source HG:
\begin{equation}
gs_i = \frac{\exp\left(\sigma\left(\mathbf{W}_{g}\left[\mathbf{h}_{G_i}||\mathbf{h}_{G_t}\right]\right)\right)}{\sum_{j=1}^{|\mathcal{G}_\emph{src}|}\exp\left(\sigma\left(\mathbf{W}_{g}\left[\mathbf{h}_{G_j}||\mathbf{h}_{G_t}\right]\right)\right)},
\end{equation}
where $\mathbf{W}_{g}$ is the trainable parameter.

\noindent\textbf{Task-level Score Submodule.}
This submodule is proposed to evaluate the consistency of each task. In meta-learning, each task consists of a support set and a query set. The support set is utilized to adapt the model to a specific task, while the query set is employed to evaluate the model's performance on unseen examples from the same task. When it comes to few-shot learning on HGs, there may exist meta-pattern differences between nodes in the support and query sets, even if they are sampled from the same class. These discrepancies in meta-patterns can impact the feedback provided by the query set, potentially resulting in inaccurate evaluations of the model's ability to acquire knowledge from the support set. However, existing methods solely consider the graph structure to assess the task importance \cite{zhang2022hg, song2022bi}, disregarding the influence of meta-pattern differences. Therefore, we propose to assess the importance of each task by examining the consistency between the meta-patterns surrounding nodes in the query set and support set. Similar to the mean-view vector $\mathbf{h}_v^\emph{mean}$ used for comparing meta-pattern distributions at the graph-level score, we utilize the sum-view vector $\mathbf{h}_v^\emph{sum}$ to represent all meta-patterns surrounding $v$ and generate representations for the support set and query set:
\begin{gather}
\mathbf{h}_{\tau^i_\emph{spt}} = \emph{mean}(\{\mathbf{h}_v^\emph{sum}|v\in\tau^i_\emph{spt}\}),
 \\
\mathbf{h}_{\tau^i_\emph{qry}} = \emph{mean}(\{\mathbf{h}_v^\emph{sum}|v\in\tau^i_\emph{qry}\}),
\end{gather}
where $\tau^i = \{\tau^i_\emph{spt}, \tau^i_\emph{qry}\}$ is a single task. Then, we concatenate the representations of the support set and query set and apply the softmax function to compute the task-level score for each task:
\begin{equation}
\emph{ts}_i = \frac{\exp\left(\sigma\left(\mathbf{W}_{t}\left[\mathbf{h}_{\tau^i_\emph{spt}}||\mathbf{h}_{\tau^i_\emph{qry}}\right]\right)\right)}{\sum_{j=1}^{|T|}\exp\left(\sigma\left(\mathbf{W}_{t}\left[\mathbf{h}_{\tau^j_\emph{spt}}||\mathbf{h}_{\tau^j_\emph{qry}}\right]\right)\right)},
\end{equation}
where $\mathbf{W}_{t}$ is the trainable parameter. Since we only compare tasks within the same HG, $T$ is the set of tasks of a single source HG. 

\noindent\textbf{Node-level Score Submodule.}
This submodule is proposed to evaluate the informativeness of each labeled node. We consider that the informativeness of the node is highly correlated to the importance of its meta-pattern instances. Consequently, we first calculate the importance of each instance as follows (here we take ${i}$-th SIP instance of node $v$, \textit{i.e.}, ${I}_{v\emph{-}i}^{\emph{SIP}}$, as an example):
\begin{equation}
\label{att_ins}    
\epsilon^{\emph{SIP}}_{v\emph{-}i} = \frac{\exp\left(\emph{LeakyReLU}\left(\mathbf{a}_\emph{SIP}^{\mathsf{T}}\cdot\left[{{s}}_{v} \| {{s}}_{v\emph{-}i}^{\emph{SIP}}\right] \right)\right)}{\sum\nolimits^{|\mathcal{I}^{\emph{SIP}}_{v}|}_{{j}=1}\exp\left(\emph{LeakyReLU}\left(\mathbf{a}_\emph{SIP}^{\mathsf{T}}\cdot\left[{{s}}_{v} \| {{s}}_{v\emph{-}j}^{\emph{SIP}}\right] \right)\right)},
\end{equation}
where $\mathcal{I}^{\emph{SIP}}_{v}$ is the set of instances that start from ${v}$ and match patterns in $\mathcal{P}^\emph{SIP}$, $\mathbf{a}_\emph{SIP}$ is the trainable attention parameter shared by SIP instances. ${{s}}_{v}$ and ${{s}}_{v\emph{-}i}^{\emph{SIP}}$ denote the initial scores of ${v}$ and $I_{v\emph{-}i}^{\emph{SIP}}$, respectively, which are computed as follows:
\begin{gather}
{{s}}_{v} = \tanh\left(\mathbf{W}_{s}\cdot \mathbf{{x}}_{v}\right),
 \\
{{s}}_{v\emph{-}i}^{\emph{SIP}} = \tanh\left(\mathbf{W}_\emph{SIP}\cdot {f}_\emph{agg}\left(\left\{\mathbf{{x}}_{u}|{u}\in{I}_{v\emph{-}i}^{\emph{SIP}}\right\}\right)\right),
\end{gather}
where $\mathbf{W}_{s}$ and $\mathbf{W}_\emph{SIP}$ are trainable parameters. The function ${f}_\emph{agg}(\cdot)$ is used to aggregate the information of nodes in each instance, such as operations like ${mean}(\cdot)$, ${max}(\cdot)$, or ${sum}(\cdot)$. Then, we utilize $|\mathcal{I}^{\emph{SIP}}_{v}|$ to quantify the popularity of node ${v}$, and then apply a sigmoid non-linearity to calculate the node-level score for the SIP category, which can be expressed as follows:
\begin{equation}
{ns}_{v}^{\emph{SIP}}={sigmoid}\left(\log\left(|\mathcal{I}^{\emph{SIP}}_{v}|+\eta\right)\cdot\sum\nolimits^{|\mathcal{I}^{\emph{SIP}}_{v}|}_{{i}=1} \epsilon^{\emph{SIP}}_{{v\emph{-}i}} \cdot {{s}}_{v\emph{-}i}^{\emph{SIP}}\right),
\end{equation}
where $\eta$ is a small constant. Next, we calculate the average score for all meta-pattern categories to obtain the node-level score:
\begin{equation}
    {ns}_{v} = ({ns}_{v}^{\emph{SAP}} + {ns}_{v}^{\emph{WAP}} + {ns}_{v}^{\emph{SIP}} + {ns}_{v}^{\emph{WIP}})/4.
\end{equation}
\noindent\textbf{Prototypical Network.}
After obtaining the node-level score for each node sample, CGFL follows the idea of prototypical networks \cite{snell2017prototypical} and calculates the weighted average of ${K}$-shot embedded nodes belonging to class ${c}$ to obtain the prototype of that class:
\begin{equation}\label{proto_fun}
\mathbf{proto}_{c} = \sum\nolimits^{K}_{{i}=1}{sc}_{i}\cdot \mathbf{h}_{i},
\end{equation}
where ${sc}_{i}$=${ns}_{i}/\sum\nolimits^{K}_{{i}=1}{ns}_{i}$ is the normalized node-level score, $\mathbf{h}_{i}$ is the embedding of node $i$ that is output by multi-view learning. The node-level score submodule is trained through meta-training and used to compute the class prototype in meta-testing.

\noindent\textbf{Loss Function.} 
During meta-training, the prototype for each class is computed by utilizing the nodes in the support set $\tau^\emph{spt}$. To determine the class of a query node $u$ in the query set $\tau^\emph{qry}$, we calculate the probability for each class based on the distance between the node embedding $\mathbf{h}_{u}$ and each prototype:
\begin{equation}\label{class_vec}
\emph{prob}({c}|{u})=\frac{\exp \left(-{d}(\mathbf{h}_{u},\mathbf{proto}_{c})\right)}{\sum_{c'} \exp \left(-{d}(\mathbf{h}_{u},\mathbf{proto}_{c'})\right)},
\end{equation}
where ${d}(\cdot)$ is a distance metric function and we adopt squared Euclidean distance \cite{snell2017prototypical}. Under the episodic training framework, the objective of each meta-training task is to minimize the classification loss between the predictions of the query set and the ground truth. The training loss of a single task $\tau$ can be defined as the average negative log-likelihood probability of assigning correct class labels:
\begin{equation}\label{loss_task}
\mathcal{L}_\tau = -\frac{1}{{N}\times {K}}\sum\nolimits_{{i}=1}^{{N}\times {K}} \log (\emph{prob}({y}^*_{i}|{v}_{i})),
\end{equation}
where ${y}^*_{i}$ is the ground truth label of ${v}_{i}$. Then, by incoporating the graph-level score and task-level score, the total meta-training loss can be defined as:
\begin{equation}\label{loss_meta}
\mathcal{L}_\emph{meta} = \sum\nolimits_{i=1}^{|\mathcal{G}_\emph{src}|}\sum\nolimits_{j=1}^{|T_i|}gs_{i}\cdot ts_{j}\cdot \mathcal{L}^i_{\tau_j},
\end{equation}
where $\mathcal{L}^i_{\tau_j}$ denotes the training loss on $j$-th task of $i$-th source HG.

\subsection{Efficiency Analysis}
Targeting the novel and complex cross-heterogeneity scenario where there may exist multiple large-scale source HGs with various node types, CGFL is efficient due to the following properties:

\noindent\textbf{Few-shot learning based on local subgraphs:} 
CGFL extracts subgraphs around $N$-way $K$-shot labeled nodes for learning, rather than processing the entire graph structure. This approach ensures the input graph size for CGFL remains small and consistent during computation, resulting in stable memory consumption and improved computational efficiency. 
 
\noindent\textbf{Independent of node type for heterogeneous information extraction:} CGFL does not rely on specific node types to capture heterogeneous information; instead, it focuses on capturing four widely existing meta-pattern categories across HGs. This ensures that the computational complexity of CGFL remains unaffected when the number of node types increases.


\section{Experiments and Analysis}
We conduct extensive experiments on four real-world datasets to answer the following research questions: \textbf{RQ1:} How does CGFL perform compared with representative and state-of-the-art methods? \textbf{RQ2:} How do the three proposed modules (\textit{i.e.}, meta-pattern extraction module, multi-view learning module, and meta-learning module) contribute to the overall performance? \textbf{RQ3:} How does the number of source HGs affect the performance of HGFL? \textbf{RQ4:} How does CGFL perform under different parameter settings?

\subsection{Experimental Settings}
\noindent\textbf{Datasets.} 
We adopt four real-world datasets: DBLP, IMDB, YELP, and PubMed. These datasets are publicly available and have been widely used in studies of node classification in HGs \cite{wang2019heterogeneous,shi2020rhine}. Details of the datasets are provided in Table \ref{dataset}. 

\begin{table}[t]
\centering
\footnotesize
\caption{Statistics of Datasets.}
\label{dataset}
\begin{tabular}{c|cc}
\toprule[1pt]
\multirow{2}{*}{\textbf{Datasets}} & \textbf{Node Type}                               & \textbf{Labeled Node Type} \\
                          & \#\textbf{Nodes}                                 & \#\textbf{Classes}         \\\midrule[0.5pt]
\multirow{2}{*}{DBLP}     & Paper/Author/Venue                      & Paper             \\
                          & 4,057 / 14,328 / 20                         & 4                 \\\midrule[0.5pt]
\multirow{2}{*}{IMDB}     & Movie/Director/Actor                    & Movie             \\
                          & 4,661 / 2,270 / 5,841                       & 4                 \\\midrule[0.5pt]
\multirow{2}{*}{YELP}     &  Business/User/Service/Star/Reservation & Business          \\
                          & 2,614 / 1,286 / 9 / 2 / 2                       & 3                 \\\midrule[0.5pt]
\multirow{2}{*}{PubMed}   & Disease/Gene/Chemical/Species           & Disease           \\
                          & 20,163 / 13,561 / 26,522 / 2,863              & 8       \\\bottomrule[1pt]      
\end{tabular}
\end{table}

\noindent\textbf{Meta-learning Settings.} 
Since the node types in any two of the four datasets are completely different, we choose to use one dataset as the target HG for meta-testing, while the remaining three datasets are used as source HGs for meta-training. This setting yields the most challenging scenario for heterogeneous graph few-shot learning, wherein the datasets for meta-training and meta-testing have entirely different heterogeneities. We repeat such challenging scenarios to conduct all four separate experiments, where each of the four datasets serves as the target HG in one experiment.

\noindent\textbf{Baselines.} 
Given the absence of dedicated solutions designed for the cross-heterogeneity graph few-shot learning problem, we select 12 representative and state-of-the-art methods as baselines. These baselines can be categorized into four groups:
\begin{itemize}[leftmargin=*]
    \item \textbf{Homogeneous GNNs}: \emph{GAT} \cite{velickovic2017graph}, \emph{SGC} \cite{wu2019simplifying} and \emph{GIN} \cite{xu2018powerful}.
    \noindent\item \textbf{Heterogeneous GNNs}: \emph{HAN} \cite{wang2019heterogeneous}, \emph{MAGNN} \cite{fu2020magnn} and \emph{SGAT} \cite{lee2022sgat}.
    \item \textbf{Few-shot learning methods}: \emph{MAML} \cite{finn2017model} and \emph{ProtoNet} \cite{snell2017prototypical}.
    \item \textbf{Graph few-shot learning methods}: \emph{Meta-GNN} \cite{zhou2019meta}, \emph{GPN} \cite{ding2020graph}, \emph{G-Meta} \cite{huang2020graph} and \emph{TENT} \cite{wang2022task}. 
\end{itemize}
While most baselines can follow our meta-learning setting with minor modifications, it is worth mentioning that HAN and MAGNN employ specialized modules to capture information from specific node types and meta-paths. Consequently, these models can only deal with HGs with single-heterogeneity. Therefore, we only train HAN and MAGNN on the target HG to evaluate the effectiveness of these methods in leveraging heterogeneous information of the target HG to address the few-shot problem.


\begin{table}[]
\centering
\caption{Node classification accuracy on four datasets.}
\label{exp_acc}
\resizebox{82mm}{51mm}{
\setlength{\tabcolsep}{1.2mm}{
\begin{tabular}{cllllllll}
\specialrule{0.1em}{2pt}{2pt}
\multicolumn{1}{c|}{}            & \multicolumn{2}{c|}{\textbf{DBLP}}                                        & \multicolumn{2}{c|}{\textbf{IMDB}}                                        & \multicolumn{2}{c|}{\textbf{YELP}}                                        & \multicolumn{2}{c}{\textbf{PubMed}}                            \\
\multicolumn{1}{c|}{}            & \multicolumn{1}{c}{\textbf{2-way}} & \multicolumn{1}{c|}{\textbf{3-way}}           & \multicolumn{1}{c}{\textbf{2-way}} & \multicolumn{1}{c|}{\textbf{3-way}}           & \multicolumn{1}{c}{\textbf{2-way}} & \multicolumn{1}{c|}{\textbf{3-way}}           & \multicolumn{1}{c}{\textbf{2-way}} & \multicolumn{1}{c}{\textbf{3-way}} \\
\specialrule{0.05em}{1.5pt}{1.5pt}
\multicolumn{9}{c}{\cellcolor[HTML]{E5E5E5}1-$\textbf{\emph{shot}}$}                        \\ \specialrule{0.05em}{1.5pt}{1.5pt}
\multicolumn{1}{c|}{GAT}         & 0.7814                    & \multicolumn{1}{l|}{0.6443}          & 0.5092                    & \multicolumn{1}{l|}{0.3421}          & 0.5839                    & \multicolumn{1}{l|}{0.3938}          & 0.5023                    & 0.3318                    \\
\multicolumn{1}{c|}{SGC}         & 0.7745                    & \multicolumn{1}{l|}{0.6599}          & 0.5142                    & \multicolumn{1}{l|}{0.3423}          & 0.5964                    & \multicolumn{1}{l|}{0.4039}          & 0.5072                    & 0.3360                    \\
\multicolumn{1}{c|}{GIN}         & 0.7804                    & \multicolumn{1}{l|}{0.6503}          & 0.5173                    & \multicolumn{1}{l|}{0.3499}          & 0.5902                    & \multicolumn{1}{l|}{0.4172}          & 0.4961                    & 0.3393                    \\\specialrule{0.05em}{1.5pt}{1.5pt}
\multicolumn{1}{c|}{HAN}         & 0.6828                    & \multicolumn{1}{l|}{0.5497}          & 0.5153                    & \multicolumn{1}{l|}{0.3514}          & 0.5681                    & \multicolumn{1}{l|}{0.3896}          & 0.5080                    & 0.3367                    \\
\multicolumn{1}{c|}{MAGNN}       & 0.7006                    & \multicolumn{1}{l|}{0.5552}          & 0.5144                    & \multicolumn{1}{l|}{0.3505}          & 0.5774                    & \multicolumn{1}{l|}{0.3935}          & 0.5092                    & 0.3375                    \\
\multicolumn{1}{c|}{SGAT}        & 0.7897                    & \multicolumn{1}{l|}{0.6631}          & 0.5234                    & \multicolumn{1}{l|}{0.3584}          & 0.5932                    & \multicolumn{1}{l|}{0.4132}          & 0.5066                    & 0.3419                    \\\specialrule{0.05em}{1.5pt}{1.5pt}
\multicolumn{1}{c|}{MAML}        & 0.6263                    & \multicolumn{1}{l|}{0.4651}          & 0.5111                    & \multicolumn{1}{l|}{0.3404}          & 0.5303                    & \multicolumn{1}{l|}{0.3674}          & 0.5071                    & 0.3232                    \\
\multicolumn{1}{c|}{ProtoNet}    & 0.6404                    & \multicolumn{1}{l|}{0.4917}          & 0.5133                    & \multicolumn{1}{l|}{0.3397}          & 0.5436                    & \multicolumn{1}{l|}{0.3665}          & 0.5037                    & 0.3242                    \\\specialrule{0.05em}{1.5pt}{1.5pt}
\multicolumn{1}{c|}{Meta-GNN}    & 0.6974                    & \multicolumn{1}{l|}{0.5423}          & 0.5175                    & \multicolumn{1}{l|}{0.3495}          & 0.5769                    & \multicolumn{1}{l|}{0.3717}          & 0.5106                    & 0.3449                    \\
\multicolumn{1}{c|}{GPN}         & 0.7611                    & \multicolumn{1}{l|}{0.6506}          & 0.5197                    & \multicolumn{1}{l|}{0.3603*}         & 0.6207                    & \multicolumn{1}{l|}{0.4266*}         & 0.5048                    & 0.3303                    \\
\multicolumn{1}{c|}{G-Meta}      & 0.7827                    & \multicolumn{1}{l|}{0.6694}          & 0.5241*                   & \multicolumn{1}{l|}{0.3573}          & 0.6274                    & \multicolumn{1}{l|}{0.4242}          & 0.5155*                   & 0.3461*                   \\
\multicolumn{1}{c|}{TENT}        & 0.8124*                   & \multicolumn{1}{l|}{0.7035*}         & 0.5174                    & \multicolumn{1}{l|}{0.3592}          & 0.6440*                   & \multicolumn{1}{l|}{0.4157}          & 0.5121                    & 0.3452                    \\\specialrule{0.05em}{1.5pt}{1.5pt}
\multicolumn{1}{c|}{CGFL}        & {$\textbf{0.8712}^1$}           & \multicolumn{1}{l|}{\textbf{0.7849}} & \textbf{0.5331}           & \multicolumn{1}{l|}{\textbf{0.3846}} & \textbf{0.7028}           & \multicolumn{1}{l|}{\textbf{0.4377}} & \textbf{0.5335}           & \textbf{0.3676}           \\
\multicolumn{1}{c|}{Improvement$^2$} & 6.75\%                    & \multicolumn{1}{l|}{10.37\%}         & 1.69\%                    & \multicolumn{1}{l|}{6.32\%}          & 8.37\%                    & \multicolumn{1}{l|}{2.54\%}          & 3.37\%                    & 5.85\%                    \\
\specialrule{0.05em}{1.5pt}{1.5pt}
\multicolumn{9}{c}{\cellcolor[HTML]{E5E5E5}3-$\textbf{\emph{shot}}$}                        \\ \specialrule{0.05em}{1.5pt}{1.5pt} 
\multicolumn{1}{c|}{GAT}         & 0.8371                    & \multicolumn{1}{l|}{0.7284}          & 0.5313                    & \multicolumn{1}{l|}{0.3669}          & 0.6393                    & \multicolumn{1}{l|}{0.4373}          & 0.4683                    & 0.3040                    \\
\multicolumn{1}{c|}{SGC}         & 0.8150                    & \multicolumn{1}{l|}{0.7389}          & 0.5377                    & \multicolumn{1}{l|}{0.3764}          & 0.6580                    & \multicolumn{1}{l|}{0.4391}          & 0.4840                    & 0.3027                    \\
\multicolumn{1}{c|}{GIN}         & 0.8433                    & \multicolumn{1}{l|}{0.7411}          & 0.5287                    & \multicolumn{1}{l|}{0.3769}          & 0.6547                    & \multicolumn{1}{l|}{0.4082}          & 0.4923                    & 0.2978                    \\\specialrule{0.05em}{1.5pt}{1.5pt}
\multicolumn{1}{c|}{HAN}         & 0.7274                    & \multicolumn{1}{l|}{0.6061}          & 0.5344                    & \multicolumn{1}{l|}{0.3764}          & 0.6796                    & \multicolumn{1}{l|}{0.4596}          & 0.5116                    & 0.3527                    \\
\multicolumn{1}{c|}{MAGNN}       & 0.7522                    & \multicolumn{1}{l|}{0.6456}          & 0.5275                    & \multicolumn{1}{l|}{0.3905}          & 0.6389                    & \multicolumn{1}{l|}{0.4359}          & 0.5266                    & 0.3789                    \\
\multicolumn{1}{c|}{SGAT}        & 0.8382                    & \multicolumn{1}{l|}{0.7426}          & 0.5332                    & \multicolumn{1}{l|}{0.3824}          & 0.6434                    & \multicolumn{1}{l|}{0.4521}          & 0.5379                    & 0.3824                    \\\specialrule{0.05em}{1.5pt}{1.5pt}
\multicolumn{1}{c|}{MAML}        & 0.6685                    & \multicolumn{1}{l|}{0.5082}          & 0.5253                    & \multicolumn{1}{l|}{0.3507}          & 0.5621                    & \multicolumn{1}{l|}{0.3967}          & 0.4997                    & 0.3237                    \\
\multicolumn{1}{c|}{ProtoNet}    & 0.6953                    & \multicolumn{1}{l|}{0.5799}          & 0.5194                    & \multicolumn{1}{l|}{0.3602}          & 0.5785                    & \multicolumn{1}{l|}{0.4231}          & 0.4927                    & 0.3117                    \\\specialrule{0.05em}{1.5pt}{1.5pt}
\multicolumn{1}{c|}{Meta-GNN}    & 0.7983                    & \multicolumn{1}{l|}{0.6953}          & 0.5309                    & \multicolumn{1}{l|}{0.3982}          & 0.6271                    & \multicolumn{1}{l|}{0.4326}          & 0.5140                    & 0.3890                    \\
\multicolumn{1}{c|}{GPN}         & 0.8450                    & \multicolumn{1}{l|}{0.7982}          & 0.5467*                   & \multicolumn{1}{l|}{0.4042*}         & 0.6766                    & \multicolumn{1}{l|}{0.4596}          & 0.5213                    & 0.3979*                   \\
\multicolumn{1}{c|}{G-Meta}      & 0.8680*                   & \multicolumn{1}{l|}{0.8138*}         & 0.5350                    & \multicolumn{1}{l|}{0.4021}          & 0.6836*                   & \multicolumn{1}{l|}{0.4577}          & 0.5392*                   & 0.3744                    \\
\multicolumn{1}{c|}{TENT}        & 0.8576                    & \multicolumn{1}{l|}{0.7538}          & 0.5212                    & \multicolumn{1}{l|}{0.3677}          & 0.6744                    & \multicolumn{1}{l|}{0.4680*}         & 0.5317                    & 0.3968                    \\\specialrule{0.05em}{1.5pt}{1.5pt}
\multicolumn{1}{c|}{CGFL}        & \textbf{0.9026}           & \multicolumn{1}{l|}{\textbf{0.8572}} & \textbf{0.5745}           & \multicolumn{1}{l|}{\textbf{0.4181}} & \textbf{0.7532}           & \multicolumn{1}{l|}{\textbf{0.4904}} & \textbf{0.5833}           & \textbf{0.4204}           \\
\multicolumn{1}{c|}{Improvement$^2$} & 3.83\%                    & \multicolumn{1}{l|}{5.06\%}          & 4.84\%                    & \multicolumn{1}{l|}{3.32\%}          & 9.24\%                    & \multicolumn{1}{l|}{4.57\%}          & 7.56\%                    & 5.35\%                                     \\\specialrule{0.1em}{2pt}{2pt}
\end{tabular}}}
\footnotesize{\leftline{* Result of the best-performing baseline.}}
\footnotesize{\leftline{$^1$ Bold numbers are the results of the best-performing method.}}
\footnotesize{\leftline{$^2$ Improvement of our CGFL over the best-performing baseline.}}
\end{table}

\noindent\textbf{Parameter Settings.} In the ${N}$-way ${K}$-shot setting, ${N}$ is set to \{2, 3\} and $\emph{K}$ is set to \{1, 3, 5\}. The number of task $\emph{m}$ is set to 100 for all datasets. To ensure fair comparisons, the embedding dimension is set to 64 for both the baselines and CGFL; the parameters for each baseline are initially set to the values reported in the original paper and then optimized through grid-searching to achieve the best performance. For CGFL, in the meta-pattern extraction module, ${N}_\emph{path}$ is set to 20, ${l}$ is set to 40, ${K}_\emph{mp}$ is set to 10, $\theta_\emph{lp}$ is set to 3. In the multi-view learning module, ${N}_\emph{mean}$ is set to 5, $\emph{f}_\theta^\emph{SAP}(\cdot)$, $\emph{f}_\theta^\emph{WAP}(\cdot)$, $\emph{f}_\theta^\emph{SIP}(\cdot)$ and $\emph{f}_\theta^\emph{WIP}(\cdot)$ are set as $\emph{mean-pooling}(\cdot)$, $\emph{K}_\emph{att}$ is set to 4, the activation $\sigma(\cdot)$ is set to $\emph{relu}(\cdot)$. In the meta-learning module, $f_\emph{agg}(\cdot)$ is set to $\emph{sum}(\cdot)$. 

\noindent\textbf{Evaluation Metrics.} The performance of all methods is evaluated by two widely used node classification metrics: \emph{Accuracy} and \emph{Macro-F1 score} \cite{velickovic2017graph,zhou2019meta}. To ensure a fair and accurate assessment of the performance of all methods, we perform 10 independent runs for each $\emph{N}$-way $\emph{K}$-shot setting and report the average results. 

\begin{table}[]
\centering
\caption{Node classification F1-score on four datasets.}
\label{exp_f1}
\resizebox{82mm}{51mm}{
\setlength{\tabcolsep}{1.2mm}{
\begin{tabular}{cllllllll}
\specialrule{0.1em}{2pt}{2pt}
\multicolumn{1}{c|}{}            & \multicolumn{2}{c|}{\textbf{DBLP}}                                        & \multicolumn{2}{c|}{\textbf{IMDB}}                                        & \multicolumn{2}{c|}{\textbf{YELP}}                                        & \multicolumn{2}{c}{\textbf{PubMed}}                            \\
\multicolumn{1}{c|}{}            & \multicolumn{1}{c}{\textbf{2-way}} & \multicolumn{1}{c|}{\textbf{3-way}}           & \multicolumn{1}{c}{\textbf{2-way}} & \multicolumn{1}{c|}{\textbf{3-way}}           & \multicolumn{1}{c}{\textbf{2-way}} & \multicolumn{1}{c|}{\textbf{3-way}}           & \multicolumn{1}{c}{\textbf{2-way}} & \multicolumn{1}{c}{\textbf{3-way}} \\
\specialrule{0.05em}{1.5pt}{1.5pt}
\multicolumn{9}{c}{\cellcolor[HTML]{E5E5E5}1-$\textbf{\emph{shot}}$}                        \\ \specialrule{0.05em}{1.5pt}{1.5pt}
\multicolumn{1}{c|}{GAT}         & 0.7634                    & \multicolumn{1}{l|}{0.6238}          & 0.4874                    & \multicolumn{1}{l|}{0.3230}          & 0.5717                    & \multicolumn{1}{l|}{0.3656}          & 0.4697                    & 0.3058                    \\
\multicolumn{1}{c|}{SGC}         & 0.7516                    & \multicolumn{1}{l|}{0.6316}          & 0.4609                    & \multicolumn{1}{l|}{0.3077}          & 0.5496                    & \multicolumn{1}{l|}{0.3754}          & 0.4451                    & 0.2757                    \\
\multicolumn{1}{c|}{GIN}         & 0.7626                    & \multicolumn{1}{l|}{0.6213}          & 0.4838                    & \multicolumn{1}{l|}{0.3218}          & 0.5705                    & \multicolumn{1}{l|}{0.3775}          & 0.4544                    & 0.3014                    \\\specialrule{0.05em}{1.5pt}{1.5pt}
\multicolumn{1}{c|}{HAN}         & 0.6569                    & \multicolumn{1}{l|}{0.5258}          & 0.4994                    & \multicolumn{1}{l|}{0.3367}          & 0.5086                    & \multicolumn{1}{l|}{0.3353}          & 0.4745                    & 0.2958                    \\
\multicolumn{1}{c|}{MAGNN}       & 0.6967                    & \multicolumn{1}{l|}{0.5492}          & 0.4851                    & \multicolumn{1}{l|}{0.3259}          & 0.5467                    & \multicolumn{1}{l|}{0.3805}          & 0.4837                    & 0.3280                    \\
\multicolumn{1}{c|}{SGAT}        & 0.7712                    & \multicolumn{1}{l|}{0.6516}          & 0.4895                    & \multicolumn{1}{l|}{0.3365}          & 0.5719                    & \multicolumn{1}{l|}{0.3745}          & 0.4828                    & 0.3372                    \\\specialrule{0.05em}{1.5pt}{1.5pt}
\multicolumn{1}{c|}{MAML}        & 0.6233                    & \multicolumn{1}{l|}{0.4622}          & 0.5088                    & \multicolumn{1}{l|}{0.3373}          & 0.5281                    & \multicolumn{1}{l|}{0.3449}          & 0.4698                    & 0.3202                    \\
\multicolumn{1}{c|}{ProtoNet}    & 0.6455                    & \multicolumn{1}{l|}{0.5343}          & 0.4831                    & \multicolumn{1}{l|}{0.3210}          & 0.5491                    & \multicolumn{1}{l|}{0.3316}          & 0.4409                    & 0.2772                    \\\specialrule{0.05em}{1.5pt}{1.5pt}
\multicolumn{1}{c|}{Meta-GNN}    & 0.6683                    & \multicolumn{1}{l|}{0.4851}          & 0.5063                    & \multicolumn{1}{l|}{0.3397}          & 0.5531                    & \multicolumn{1}{l|}{0.3603}          & 0.5039                    & 0.3222                    \\
\multicolumn{1}{c|}{GPN}         & 0.7492                    & \multicolumn{1}{l|}{0.6294}          & 0.5147*                   & \multicolumn{1}{l|}{0.3376}          & 0.6082*                   & \multicolumn{1}{l|}{0.3941}          & 0.5008                    & 0.3447*                   \\
\multicolumn{1}{c|}{G-Meta}      & 0.7649                    & \multicolumn{1}{l|}{0.6427}          & 0.5084                    & \multicolumn{1}{l|}{0.3436*}         & 0.5439                    & \multicolumn{1}{l|}{0.3954*}         & 0.5075*                   & 0.3372                    \\
\multicolumn{1}{c|}{TENT}        & 0.8081*                   & \multicolumn{1}{l|}{0.6767*}         & 0.5029                    & \multicolumn{1}{l|}{0.3205}          & 0.5567                    & \multicolumn{1}{l|}{0.3814}          & 0.4932                    & 0.3358                    \\\specialrule{0.05em}{1.5pt}{1.5pt}
\multicolumn{1}{c|}{CGFL}        & \textbf{0.8456$^1$}           & \multicolumn{1}{l|}{\textbf{0.7536}} & \textbf{0.5244}           & \multicolumn{1}{l|}{\textbf{0.3469}} & \textbf{0.6876}           & \multicolumn{1}{l|}{\textbf{0.4126}} & \textbf{0.5155}           & \textbf{0.3520}           \\
\multicolumn{1}{c|}{Improvement$^2$} & 4.43\%                    & \multicolumn{1}{l|}{10.20\%}         & 1.85\%                    & \multicolumn{1}{l|}{0.95\%}          & 11.55\%                   & \multicolumn{1}{l|}{4.17\%}          & 1.11\%                    & 2.07\%                    \\                 
\specialrule{0.05em}{1.5pt}{1.5pt}
\multicolumn{9}{c}{\cellcolor[HTML]{E5E5E5}3-$\textbf{\emph{shot}}$}                        \\ \specialrule{0.05em}{1.5pt}{1.5pt} 
\multicolumn{1}{c|}{GAT}         & 0.8042                    & \multicolumn{1}{l|}{0.7067}          & 0.5133                    & \multicolumn{1}{l|}{0.3545}          & 0.6059                    & \multicolumn{1}{l|}{0.4105}          & 0.4493                    & 0.2742                    \\
\multicolumn{1}{c|}{SGC}         & 0.8092                    & \multicolumn{1}{l|}{0.7168}          & 0.5229                    & \multicolumn{1}{l|}{0.3649}          & 0.6388                    & \multicolumn{1}{l|}{0.4208}          & 0.4388                    & 0.2616                    \\
\multicolumn{1}{c|}{GIN}         & 0.8001                    & \multicolumn{1}{l|}{0.7282}          & 0.5180                    & \multicolumn{1}{l|}{0.3546}          & 0.6470                    & \multicolumn{1}{l|}{0.4011}          & 0.4562                    & 0.2695                    \\\specialrule{0.05em}{1.5pt}{1.5pt} 
\multicolumn{1}{c|}{HAN}         & 0.6909                    & \multicolumn{1}{l|}{0.5839}          & 0.5231                    & \multicolumn{1}{l|}{0.3659}          & 0.6574                    & \multicolumn{1}{l|}{0.4396}          & 0.4961                    & 0.3439                    \\
\multicolumn{1}{c|}{MAGNN}       & 0.7489                    & \multicolumn{1}{l|}{0.6150}          & 0.5022                    & \multicolumn{1}{l|}{0.3729}          & 0.6300                    & \multicolumn{1}{l|}{0.4230}          & 0.5093                    & 0.3668                    \\
\multicolumn{1}{c|}{SGAT}        & 0.8123                    & \multicolumn{1}{l|}{0.7146}          & 0.5153                    & \multicolumn{1}{l|}{0.3620}          & 0.6355                    & \multicolumn{1}{l|}{0.4361}          & 0.5118                    & 0.3723                    \\\specialrule{0.05em}{1.5pt}{1.5pt} 
\multicolumn{1}{c|}{MAML}        & 0.6453                    & \multicolumn{1}{l|}{0.5052}          & 0.5167                    & \multicolumn{1}{l|}{0.3466}          & 0.5598                    & \multicolumn{1}{l|}{0.3805}          & 0.4907                    & 0.3055                    \\
\multicolumn{1}{c|}{ProtoNet}    & 0.6343                    & \multicolumn{1}{l|}{0.5136}          & 0.4484                    & \multicolumn{1}{l|}{0.2717}          & 0.5217                    & \multicolumn{1}{l|}{0.3530}          & 0.4771                    & 0.3038                    \\\specialrule{0.05em}{1.5pt}{1.5pt} 
\multicolumn{1}{c|}{Meta-GNN}    & 0.7723                    & \multicolumn{1}{l|}{0.6553}          & 0.5248                    & \multicolumn{1}{l|}{0.3656}          & 0.5902                    & \multicolumn{1}{l|}{0.4207}          & 0.5076                    & 0.3715                    \\
\multicolumn{1}{c|}{GPN}         & 0.8295                    & \multicolumn{1}{l|}{0.7825*}         & 0.5316*                   & \multicolumn{1}{l|}{0.3811*}         & 0.6451                    & \multicolumn{1}{l|}{0.4338}          & 0.5185                    & 0.3797*                   \\
\multicolumn{1}{c|}{G-Meta}      & 0.8461*                   & \multicolumn{1}{l|}{0.7769}          & 0.5246                    & \multicolumn{1}{l|}{0.3559}          & 0.6494                    & \multicolumn{1}{l|}{0.4266}          & 0.5313*                   & 0.3319                    \\
\multicolumn{1}{c|}{TENT}        & 0.8396                    & \multicolumn{1}{l|}{0.7423}          & 0.5089                    & \multicolumn{1}{l|}{0.3399}          & 0.6592*                   & \multicolumn{1}{l|}{0.4430*}         & 0.5249                    & 0.3566                    \\\specialrule{0.05em}{1.5pt}{1.5pt} 
\multicolumn{1}{c|}{CGFL}        & \textbf{0.8974}           & \multicolumn{1}{l|}{\textbf{0.8512}} & \textbf{0.5539}           & \multicolumn{1}{l|}{\textbf{0.3957}} & \textbf{0.7007}           & \multicolumn{1}{l|}{\textbf{0.4664}} & \textbf{0.5615}           & \textbf{0.4013}           \\
\multicolumn{1}{c|}{Improvement$^2$} & 5.72\%                    & \multicolumn{1}{l|}{8.07\%}          & 4.03\%                    & \multicolumn{1}{l|}{3.69\%}          & 5.92\%                    & \multicolumn{1}{l|}{5.02\%}          & 5.38\%                    & 5.38\%                   
                      \\\specialrule{0.1em}{2pt}{2pt}
\end{tabular}}}
\footnotesize{\leftline{* Result of the best-performing baseline.}}
\footnotesize{\leftline{$^1$ Bold numbers are the results of the best-performing method.}}
\footnotesize{\leftline{$^2$ Improvement of our CGFL over the best-performing baseline.}}
\end{table}

\begin{figure}[t]
\centering
\scalebox{0.30}{\includegraphics{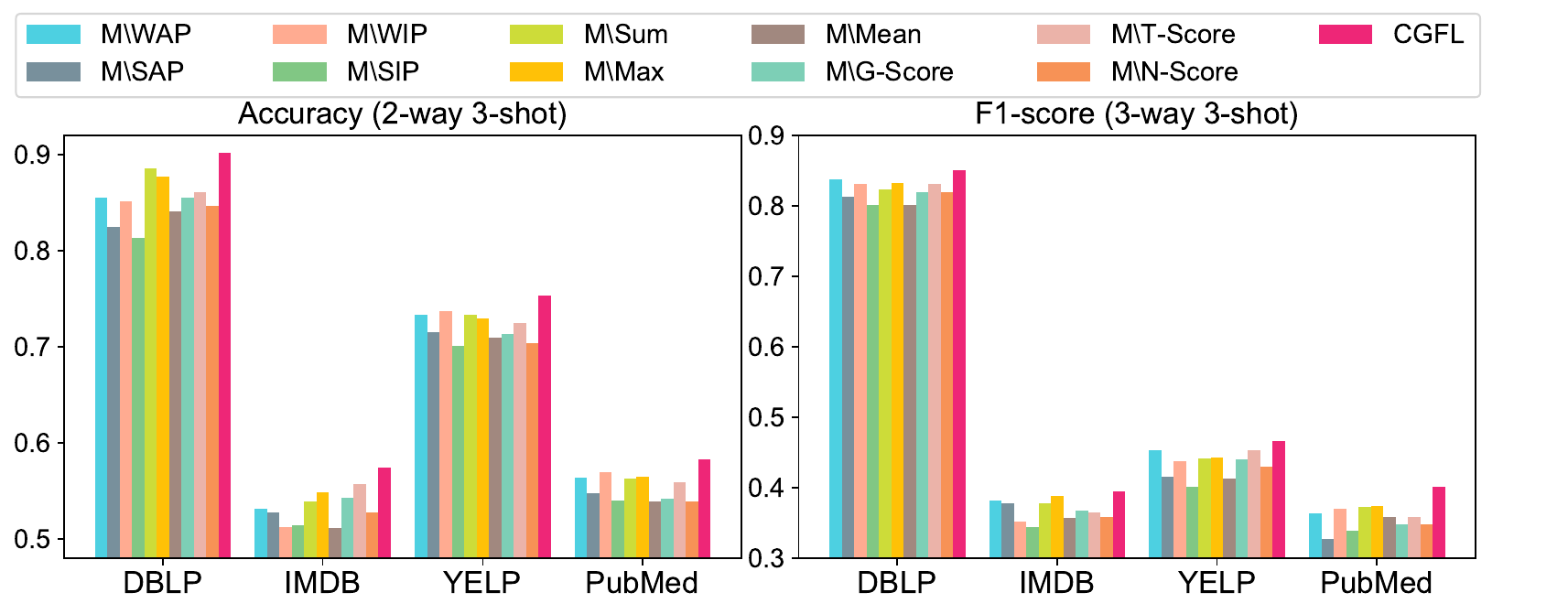}}
\caption{Node classification performance of CGFL variants.}
\label{abl}
\end{figure}

\subsection{Experimental Results}
\noindent\textbf{Performance Comparison with Baselines (RQ1).}
Tables \ref{exp_acc} and \ref{exp_f1} present the performance comparison between CGFL and baselines. The "DBLP" column presents the results of the experiment where the DBLP dataset is set as the target HG and the rest three datasets are set as source HGs. The results demonstrate a substantial improvement of CGFL over the best-performing baselines. On average, CGFL achieves a 5.56\% increase in accuracy (ranging from 1.69\% to 10.37\%) and a 4.97\% increase in F1-score (ranging from 0.95\% to 11.55\%). This improvement can be attributed to the utilization of the proposed multi-view HGNN and score module in CGFL, which effectively learn generalized knowledge across HGs. In contrast, the baselines struggle to generalize information from source HGs with different heterogeneities. thereby resulting in inferior performance when applied to a new heterogeneity with few-labeled data.

In addition, it is worth noting that some baselines exhibit a decline in performance as the number of labeled samples (\textit{i.e.}, $K$) increases. For instance, on the {PubMed} dataset, the performance of homogeneous graph methods ({GAT}, {SGC}, {GIN}) and few-shot learning methods ({MAML}, {ProtoNet}) deteriorates in the \emph{3-shot} scenario compared to the \emph{1-shot} scenario. This decline can be attributed to their limited capacity to effectively extract generalized knowledge from source HGs, leading to negative transfer. In contrast, CGFL adopts the proposed score module to evaluate the transferability of source HG data, thereby mitigating negative transfer.



\begin{table}[]
\centering
\caption{Node classification performance of CGFL with different numbers of source HGs on four datasets.}
\label{exp_nhg}
\resizebox{80mm}{32mm}
{\setlength{\tabcolsep}{1.2mm}{
\begin{tabular}{c|ll|ll|ll|ll}
\specialrule{0.1em}{2pt}{2pt}
\multicolumn{1}{c|}{\multirow{2}{*}{\begin{tabular}[c]{@{}c@{}}\#Source\\ HGs\end{tabular}}} & \multicolumn{2}{c|}{\textbf{DBLP}}                                         & \multicolumn{2}{c|}{\textbf{IMDB}}                                        & \multicolumn{2}{c|}{\textbf{YELP}}                                         & \multicolumn{2}{c}{\textbf{PubMed}}                                      \\
\multicolumn{1}{c|}{}                                                                        & \multicolumn{1}{c}{\textbf{2-way}}  & \multicolumn{1}{c|}{\textbf{3-way}}  & \multicolumn{1}{c}{\textbf{2-way}} & \multicolumn{1}{c|}{\textbf{3-way}}  & \multicolumn{1}{c}{\textbf{2-way}}  & \multicolumn{1}{c|}{\textbf{3-way}}  & \multicolumn{1}{c}{\textbf{2-way}}  & \multicolumn{1}{c}{\textbf{3-way}} \\
\specialrule{0.05em}{1.5pt}{1.5pt}
\multicolumn{9}{c}{\cellcolor[HTML]{E5E5E5}\textbf{Accuracy:} 1-$\textbf{\emph{shot}}$}                        \\ \specialrule{0.05em}{1.5pt}{1.5pt}
\multicolumn{1}{c|}{Rand-1} & 0.8382                    & 0.7305                     & 0.5084                    & 0.3382                     & 0.6512                    & 0.4042                     & 0.5013                    & 0.3346                    \\
\multicolumn{1}{c|}{Rand-2} & 0.8644                    & 0.7542                     & 0.5193                    & 0.3705                     & 0.6745                    & 0.4201                     & 0.5122                    & 0.3575                    \\                          
\multicolumn{1}{c|}{All-3}    & \textbf{0.8712}           & \textbf{0.7849}            & \textbf{0.5331}           & \textbf{0.3846}            & \textbf{0.7028}           & \textbf{0.4377}            & \textbf{0.5335}           & \textbf{0.3676}           \\
\specialrule{0.05em}{1.5pt}{1.5pt}
\multicolumn{9}{c}{\cellcolor[HTML]{E5E5E5}\textbf{Accuracy:} 3-$\textbf{\emph{shot}}$}                        \\ \specialrule{0.05em}{1.5pt}{1.5pt}  
\multicolumn{1}{c|}{Rand-1} & 0.8746                    & 0.8226                     & 0.5475                    & 0.3977                     & 0.7079                    & 0.4657                     & 0.5356                    & 0.3839                    \\
\multicolumn{1}{c|}{Rand-2} & 0.8917                    & 0.8423                     & 0.5646                    & 0.4129                     & 0.7375                    & 0.4812                     & 0.5677                    & 0.4065                    \\
\multicolumn{1}{c|}{All-3}    & \textbf{0.9026}           & \textbf{0.8572}            & \textbf{0.5745}           & \textbf{0.4181}            & \textbf{0.7532}           & \textbf{0.4904}            & \textbf{0.5833}           & \textbf{0.4204}          \\\specialrule{0.05em}{1.5pt}{1.5pt}  
\multicolumn{9}{c}{\cellcolor[HTML]{E5E5E5}\textbf{F1-score:} 1-$\textbf{\emph{shot}}$}                        \\ \specialrule{0.05em}{1.5pt}{1.5pt}
\multicolumn{1}{c|}{Rand-1} & 0.8008                    & 0.7148                     & 0.5035                    & 0.3103                     & 0.6427                    & 0.3951                     & 0.4863                    & 0.3161                    \\
\multicolumn{1}{c|}{Rand-2} & 0.8268                    & 0.7343                     & 0.5133                    & 0.3262                     & 0.6651                    & 0.4035                     & 0.5028                    & 0.3372                    \\                        
\multicolumn{1}{c|}{All-3}    & \textbf{0.8456}           & \textbf{0.7536}            & \textbf{0.5244}           & \textbf{0.3469}            & \textbf{0.6876}           & \textbf{0.4126}            & \textbf{0.5155}           & \textbf{0.3520}           \\
\specialrule{0.05em}{1.5pt}{1.5pt}
\multicolumn{9}{c}{\cellcolor[HTML]{E5E5E5}\textbf{F1-score:} 3-$\textbf{\emph{shot}}$}                        \\ \specialrule{0.05em}{1.5pt}{1.5pt}  
\multicolumn{1}{c|}{Rand-1} & 0.8622                    & 0.8149                     & 0.5253                    & 0.3546                     & 0.6754                    & 0.4243                     & 0.5223                    & 0.3736                    \\
\multicolumn{1}{c|}{Rand-2} & 0.8864                    & 0.8394                     & 0.5500                    & 0.3794                     & 0.6925                    & 0.4441                     & 0.5439                    & 0.3943                    \\
\multicolumn{1}{c|}{All-3}    & \textbf{0.8974}           & \textbf{0.8512}            & \textbf{0.5539}           & \textbf{0.3957}            & \textbf{0.7007}           & \textbf{0.4664}            & \textbf{0.5615}           & \textbf{0.4013}          \\\specialrule{0.1em}{2pt}{2pt}         
\end{tabular}}}
\end{table}

\noindent\textbf{Ablation Study (RQ2).} We create ten variants of CGFL to investigate the impact of its three main modules: (1) Four variants are created to investigate the impact of the meta-pattern extraction module by removing meta-patterns of SAP, WAP, SIP, and WIP. These variants are denoted as \textbf{M\textbackslash SAP}, \textbf{M\textbackslash WAP}, \textbf{M\textbackslash SIP}, and \textbf{M\textbackslash WIP}, respectively. (2) Three variants are developed to explore the influence of the multi-view learning module by removing the sum-view, max-view, and mean-view. These variants are denoted as \textbf{M\textbackslash Sum}, \textbf{M\textbackslash Max}, and \textbf{M\textbackslash Mean}, respectively. (3) Three variants are created to study the impact of the score module by removing submodules of the graph-level score, task-level score, and node-level score. These variants are named \textbf{M\textbackslash G-Score}, \textbf{M\textbackslash T-Score}, and \textbf{M\textbackslash N-Score}, respectively. From Fig. \ref{abl}, we have the following observations:

\begin{itemize}[leftmargin=*]
    \item {M\textbackslash SAP}, {M\textbackslash WAP}, {M\textbackslash SIP}, and {M\textbackslash WIP} all exhibit inferior performance compared to the original CGFL. This indicates that exploring each of the four meta-pattern categories is crucial for effective knowledge transfer across HGs. Notably, the removal of SAP and SIP results in a more significant decline in performance (approximate decreases of 7\% and 8\% respectively) compared to WAP and WIP (approximate decreases of 3\% and 5\% respectively). This highlights the importance of extracting SAP and SIP and emphasizes the necessity to distinguish between meta-patterns based on their relationship strength levels.
    
    \item {M\textbackslash Sum}, {M\textbackslash Max}, {M\textbackslash Mean} are outperformed by CGFL. This suggests that learning generalized relations from all three views enhances the generalizability across HGs. 
    \item {M\textbackslash G-Score}, {M\textbackslash T-Score} and {M\textbackslash N-Score} yield inferior performance compared to CGFL. This indicates that the score module plays a crucial role in evaluating source HG data and facilitating the extraction of generalized knowledge across HGs. Notably, the node-level score submodule has the greatest influence on performance, because it not only measures the informativeness of nodes in the source HGs but also determines the importance of few-labeled nodes to derive robust prototypes in the target HG. 
\end{itemize}

\noindent\textbf{Impact of the Number of Source HGs (RQ3).} We investigate whether CGFL can effectively extract generalized knowledge from different types of heterogeneities by varying the number of source HGs. The results presented in Table \ref{exp_nhg} illustrate that as the number of source HGs increases, CGFL consistently demonstrates improved performance. This is due to CGFL's capability to enrich generalized knowledge by learning from more source HGs with various heterogeneities and effectively transfer the enriched generalized knowledge to the target HG.

\begin{figure}[t]
\centering
\scalebox{0.343}{\includegraphics{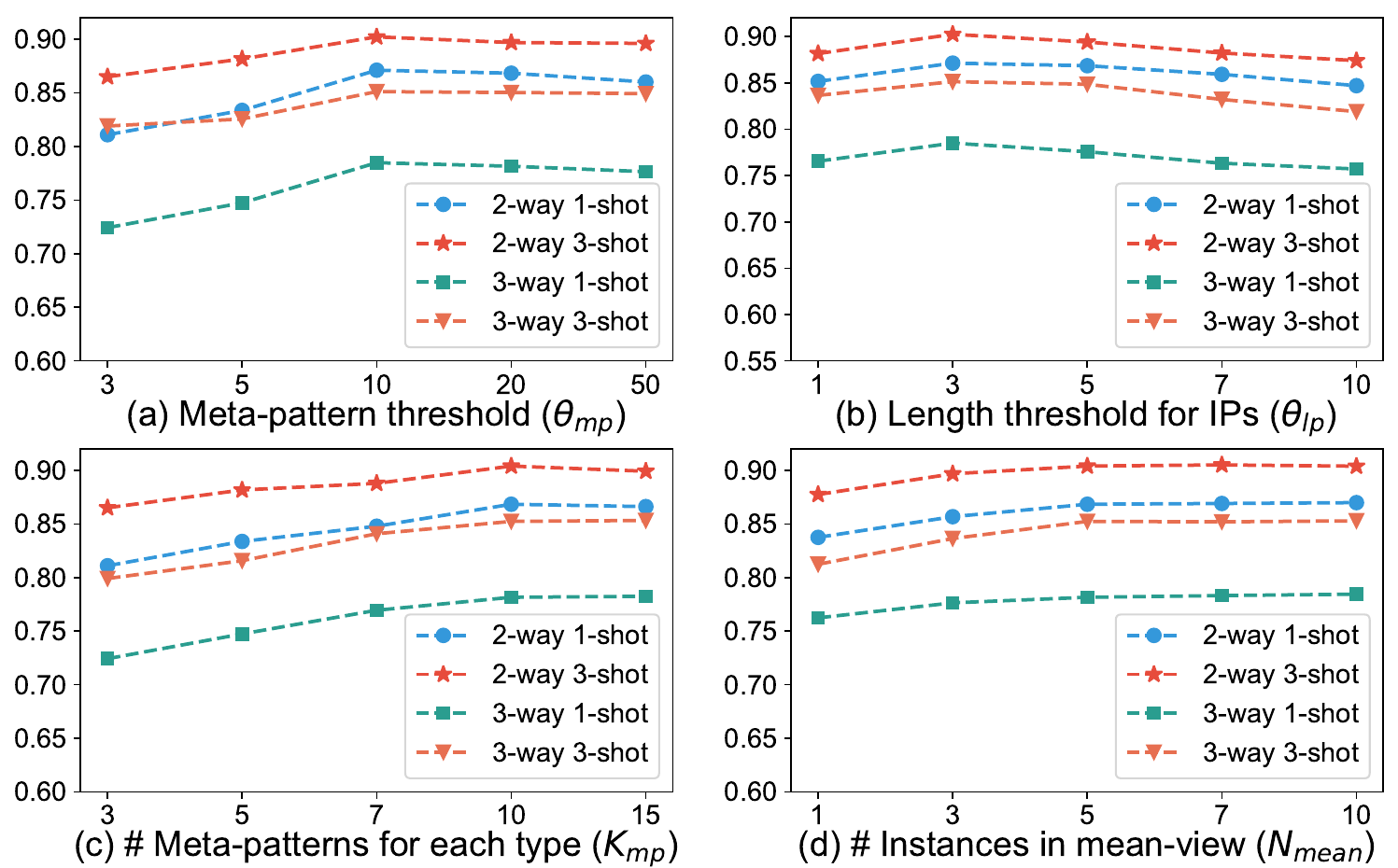}}
\caption{Node classification accuracy of CGFL with different parameter settings on DBLP dataset.}
\label{para}
\end{figure}

\noindent\textbf{Parameter Study (RQ4).} We evaluate the sensitivity of several important parameters in CGFL, and show their impacts in Fig. \ref{para}. For the meta-pattern threshold $\theta_\emph{mp}$ and length threshold $\theta_\emph{lp}$, moderate values should be set around 10 and 3 respectively. This selection helps effectively categorize meta-patterns into general categories. For the number of meta-patterns $K_\emph{mp}$, a larger value (greater than 10) generally yields better results. This is because a larger $K_\emph{mp}$ allows CGFL to capture a broader range of diverse meta-patterns. For the number of instances in the mean-view $N_\emph{mean}$, performance stabilizes when $N_\emph{mean}$ exceeds 5. This is because subsequently extracted instances may not possess distinct characteristics that contribute to the learning of meta-pattern distributions.

\section{Conclusion}
In this paper, we propose a novel cross-heterogeneity graph few-shot learning problem, and provide a solution called CGFL. CGFL incorporates a multi-view learning module to efficiently extract generalized knowledge across source HGs, and adopts a score-based meta-learning module to transfer the knowledge to the target HG with different heterogeneity for few-shot learning. Extensive experiments demonstrate the superior performance of our CGFL. In the future work, we plan to further enhance CGFL by distinguishing between heterogeneity-specific knowledge and heterogeneity-cross knowledge. Additionally, we plan to improve the meta-learning module in CGFL by considering the mutual information between source HGs and tasks originating from different source HGs. 

\bibliographystyle{ACM-Reference-Format}
\balance
\bibliography{cikm23}

\end{document}